\documentclass{article}
\usepackage{ijcai21}

\usepackage{times}
\usepackage{soul}
\usepackage{url}
\usepackage[hidelinks]{hyperref}
\usepackage[utf8]{inputenc}
\usepackage[small]{caption}
\usepackage{graphicx}
\usepackage{amsmath}
\usepackage{amsthm}
\usepackage{amssymb}
\usepackage{booktabs}
\usepackage{algorithm}
\usepackage{algorithmic}
\usepackage{multirow}
\usepackage{subfig}
\title{Rescuing Deep Hashing from Dead Bits Problem}

\author{
	Shu Zhao$^{1,2}$\and
	Dayan Wu$^1$\footnote{Contact Author}\and
	Yucan Zhou$^1$\and
	Bo Li$^1$\And
	Weiping Wang$^1$\\
	\affiliations
	$^1$Institute of Information Engineering,
	Chinese Academy of Sciences\\
	$^2$School of Cyber Security, University
	of Chinese Academy of Sciences\\
	\emails
	\{zhaoshu, wudayan, zhouyucan, libo, wangweiping\}@iie.ac.cn
}

\begin{document}

\maketitle

\begin{abstract}
	Deep hashing methods have shown great retrieval accuracy and efficiency in large-scale image retrieval. How to optimize discrete hash bits is always the focus in deep hashing methods. A common strategy in these methods is to adopt an activation function, e.g. $\operatorname{sigmoid}(\cdot)$ or $\operatorname{tanh}(\cdot)$, and minimize a quantization loss to approximate discrete values. However, this paradigm may make more and more hash bits stuck into the wrong saturated area of the activation functions and never escaped. We call this problem ``Dead Bits Problem~(DBP)''. Besides, the existing quantization loss will aggravate DBP as well. In this paper, we propose a simple but effective gradient amplifier which acts before activation functions to alleviate DBP. Moreover, we devise an error-aware quantization loss to further alleviate DBP. It avoids the negative effect of quantization loss based on the similarity between two images. The proposed gradient amplifier and error-aware quantization loss are compatible with a variety of deep hashing methods. Experimental results on three datasets demonstrate the efficiency of the proposed gradient amplifier and the error-aware quantization loss.
  
\end{abstract}

\section{Introduction}
Hashing has been widely used in image retrieval~\cite{cao2017hashnet,jiang2018asymmetric,wu2019deep,zhao2020asymmetric}, video retrieval~\cite{yuan2020central}, and cross-modal retrieval~\cite{jiang2017deep} due to its high computation efficiency and low storage cost. Traditional hashing methods are based on hand-crafted features. The representative methods include LSH~\cite{gionis1999similarity}, ITQ~\cite{gong2012iterative} and SDH~\cite{shen2015supervised}.

\begin{figure}
	\centering
	\includegraphics[width=1\columnwidth]{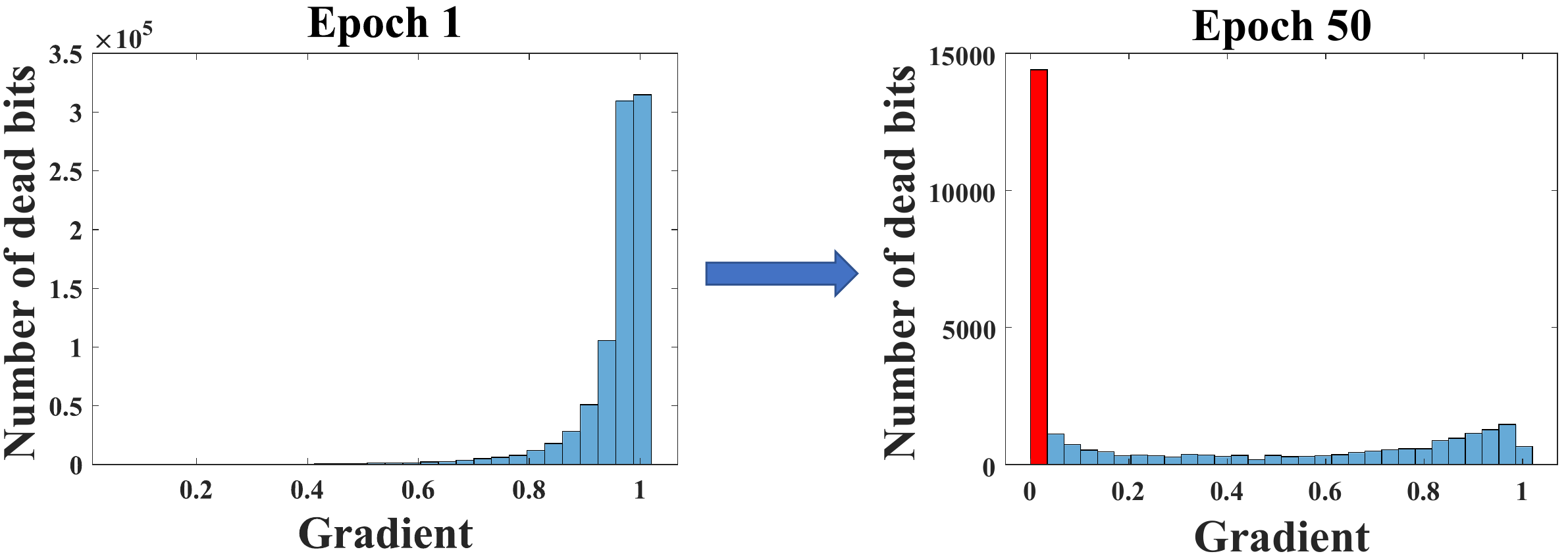}
	\caption{Existing deep hashing methods will push more and more dead bits~(red bar) into the saturated area of activation functions and these bits will never escape from it during training phase, leading to the retrieval performance degradation. Best viewed in color.}
	\label{fig:dbp}
\end{figure}

In recent years, deep hashing methods have greatly improved the retrieval performance due to their powerful representation ability. Large number of deep hashing methods have been proposed, including single-label hashing methods~\cite{gui2017fast,cao2018deep,fan20deep,zhao2020asymmetric,li2020general} and multi-label hashing methods~\cite{zhao2015deep,lai2016instance}. However, for all these deep hashing methods, the optimization is an intractable problem because of the discrete property of the binary hash codes.

\begin{figure*}
	\centering
	\includegraphics[width=2\columnwidth]{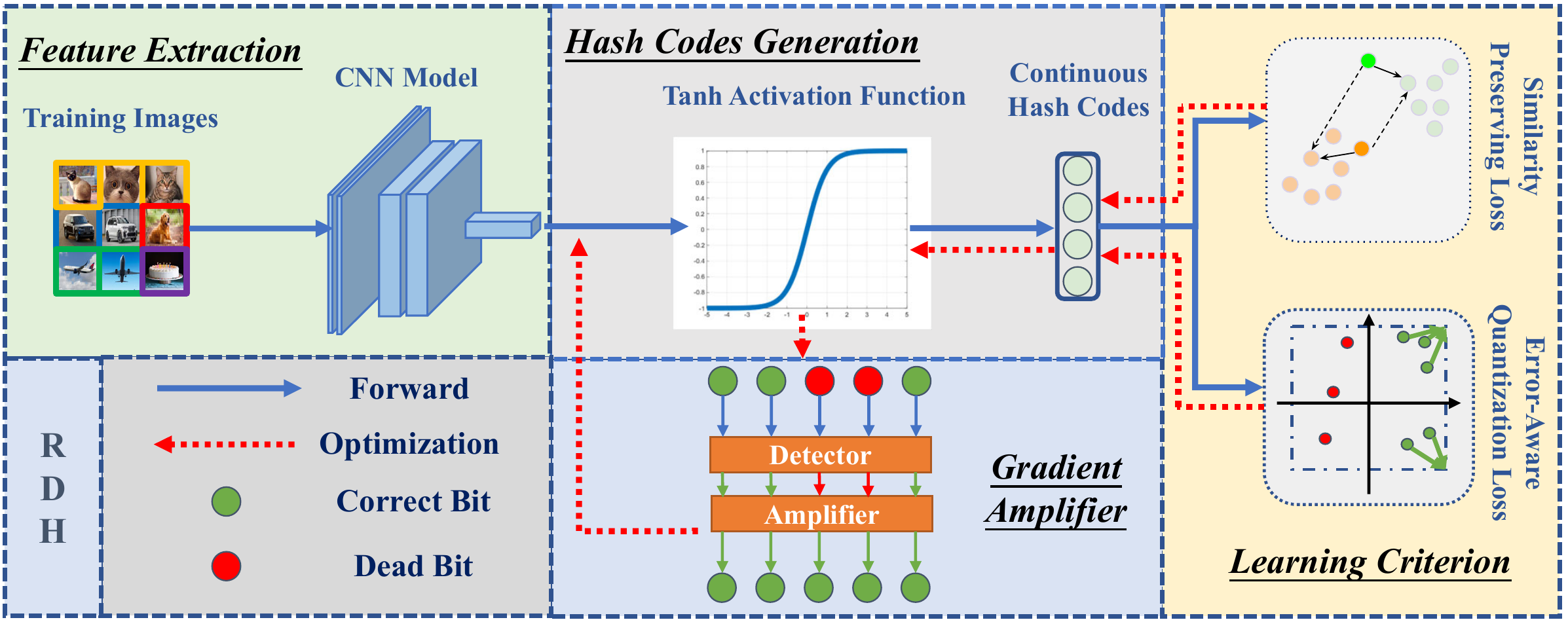}
	\caption{Framework of our proposed method. It consists of two components: a Gradient Amplifier and an Error-Aware Quantization Loss. The Gradient Amplifier aims to detect and rescue the dead bits. The Error-Aware Quantization Loss adaptively selects the bits that can be safely quantified and delegates others to the similarity preserving loss. Best viewed in color.}
	\label{fig:arch}
\end{figure*}

Previous methods usually adopt binary approximation to tackle the above problem. A common strategy in these methods is to adopt an activation function, e.g. $\operatorname{sigmoid}(\cdot)$ or $\operatorname{tanh}(\cdot)$, and minimize a quantization loss to approximate the discrete values. However, this paradigm may make more and more wrong hash bits stuck into the saturated area of the activation functions, and these wrong hash bits will never escape~(as shown in Figure~\ref{fig:dbp}). We call this problem ``Dead Bits Problem~(DBP)''. Besides, we call the wrong bits~(red bar) stuck in the saturated area as ``Dead Bits'', because the gradients are nearly zero. What's more, the number of dead bits will increase with the training epoch. This is because different bits have different change trend, i.e. \textit{uncertainty}~\cite{fu2020deep}. As a result, the direction of optimization of each hash bit is oscillatory. Some bits may ``incautiously'' fall into the saturated area and hard to escape. 

Moreover, the quantization loss widely used in hashing methods will also aggravate DBP, such as $\text{L}_2$ quantization loss~\cite{li2015feature} or $\log \cosh$ quantization loss~\cite{zhu2016deep}. These quantization loss functions drag bits into $-1$ or $+1$ simply according to their current sign, which may conflict with the similarity preserving requirement. For example, if the hash bits of two similar images fall into different sides of zero, the quantization loss will push them apart, while the similarity preserving loss will pull them together.

To address there issues, we propose a simple but effective deep hashing component named Gradient Amplifier which can detect and rescue dead bits by amplifying their gradients. Moreover, we devise an error-aware quantization loss function. For correct bits, it adaptively reduce their quantization loss to generate high-quality hash codes. For dead bits, they are propagated to the similarity preserving loss. This will avoid the negative effect brought by the original quantization loss. 

The main contributions of this work can be summarized as follows:
\begin{itemize}
	\item To the best of our knowledge, we are the first to formalize the ``Dead bits Problem'': the saturated area of activation function and existing quantization loss will ``kill'' more and more hash bits. 
	\item We propose a gradient amplifier which detects the dead bits and amplifies their gradients to rescue them. Furthermore, we design an error-aware quantization loss, which will further alleviate the DBP. The proposed gradient amplifier and error-aware quantization loss can be compatible with a various of deep hashing methods. 
	\item Extensive experiments on three datasets demonstrate the efficiency of the proposed gradient amplifier and the error-aware quantization loss.
\end{itemize}

\section{Related Work}
Learning to hash aims to project data from high-dimensional space into low-dimensional Hamming space and can be categorized into traditional hashing methods and deep hashing methods. 

\textbf{Traditional hashing methods} leverage hand-crafted features to learn hash function. LSH~\cite{gionis1999similarity} generates a random projection matrix to map the features into hash codes. ITQ~\cite{gong2012iterative} uses PCA to perform dimension reduction. SDH~\cite{shen2015supervised} adopts discrete cyclic coordinate descent~(DCC) to optimize hash codes directly.

\textbf{Deep hashing methods} involve CNN models into the learning of hash codes. CNNH~\cite{xia2014supervised} is a two-stage hashing method that utilizes CNN model to generate hash codes. DNNH~\cite{lai2015simultaneous} is the first end-to-end deep hashing method that learns features and hash codes simultaneously. DPSH~\cite{li2015feature} leverages pairwise supervised information to learn hash function. And DSDH~\cite{li2020general} introduces classification information into the training process and optimizes database codes by DCC~\cite{shen2015supervised}. Recently, asymmetric architecture of deep hashing has shown great potential to improve the retrieval performance. DAPH~\cite{shen2017deep} adopts two different CNN model to learn a hash function simultaneously. ADSH~\cite{jiang2018asymmetric} utilizes a CNN for query images, while the database codes are learned directly by DCC. DIHN~\cite{wu2019deep} aims to learn a hash function incrementally. CCDH~\cite{zhao2020asymmetric} leverages variational autoencoder to update the CNN model and database codes efficiently.

For all the deep hashing methods, the optimization for binary hash codes is remain an intractable problem. To approximate the non-differentiable $\operatorname{sign}(\cdot)$ function, $\operatorname{tanh}(\cdot)$ and quantization loss are  introduced and widely used in hashing methods~\cite{zhu2016deep,cao2017hashnet,wu2019deep,zhao2020asymmetric}. However, few research focuses at the problem leading by the saturated area of activation function and the quantization loss in pairwise learning. They will push bits toward wrong direction and can not escape from the saturated area.

\section{Preliminaries: Deep Hashing Models}
Assume we have $N$ training images denoted as $\mathbb{X} = \{\mathbf{x}_i\}_{i=1}^{N}$, and an $N \times N$ similarity matrix $\mathbf{S}$, where $\mathbf{S}_{ij} = 1$ if $\mathbf{x}_i$ and $\mathbf{x}_j$ have same label and otherwise $\mathbf{S}_{ij} = 0$. Deep hashing model aims to learn a hash function $\operatorname{F}(\mathbf{x}; \theta)$, mapping an input image $\mathbf{x}$ into a $K$-dimension embedding, where $\theta$ is the parameters of the model. Subsequently, we apply $\operatorname{sign}(\cdot)$ function to the embeddings for obtaining binary hash codes $\mathbf{B} = \{\mathbf{b}_i\}_{i=1}^{N} \in \{-1, +1\}^{N \times K}$. However, because the gradient of $\operatorname{sign}(\cdot)$ function is always $0$ expects $x = 0$, we can not back-propagate it directly. To address this issue, a common strategy in these methods is to adopt $\operatorname{sigmoid}(\cdot)$ or $\operatorname{tanh}(\cdot)$ function and minimize a quantization loss, i.e. $\text{L}_2$ quantization loss, to approximate it. Without loss of generality, we utilize the $\operatorname{tanh}(\cdot)$ to illustrate our approach. we In the next sections, we firstly characterize a common and serious problem in these saturated functions and quantization losses, then propose a simple but effective method to solve it.

\section{Dead Bits Problem}



To study DBP, we firstly investigate the gradient of hash loss functions. Without loss of generality, here we choose a representative hash loss, which is formulated as:
\begin{equation}
	\begin{aligned}
		\mathcal{L} &= \sum_{i=1}^{N}\sum_{j=1}^{N}\left(\log \left(1 + e^{\Theta_{ij}}\right) - \mathbf{S}_{ij}\Theta_{ij}\right) \\
					&+ \eta \sum_{i=1}^{N}\Vert \mathbf{h}_i -\operatorname{sign}(\mathbf{h}_i)\Vert_2^2,
	\end{aligned}
	\label{eq:1}
\end{equation}
where $\mathbf{h}_i \in [-1, 1]^K$ is the relaxed continuous codes, $\Theta_{ij} = \frac{\mathbf{h}_{i}^T \mathbf{h}_j}{2}$, $\eta$ is the hyper-parameter to control the balance between similarity loss~(first term) and quantization loss~(second term). This form of hash loss is widely used in hash methods~\cite{li2015feature,zhu2016deep,cao2017hashnet,chen2019deep,li2020general}.

Next, we calculate the derivative of Eq.~\eqref{eq:1} w.r.t. $\operatorname{F}^k(\mathbf{x}_i;\theta)$, where $k$ is the $k$-th dimension of embedding.

\begin{equation}
	\begin{aligned}
		&\frac{\partial\mathcal{L}}{\partial\operatorname{F}^k(\mathbf{x}_i;\theta)} = \frac{\partial\mathcal{L}}{\partial\mathbf{h}_i^k}\frac{\partial\mathbf{h}_i^k}{\partial\operatorname{F}^k(\mathbf{x}_i;\theta)}\\
		&\quad= \left\{\sum_{j=1}^N\left[\sigma(\Theta_{ij}) - \mathbf{S}_{ij}\right]\mathbf{h}_j^k + 2\eta\left[\mathbf{h}_i^k - \operatorname{sign}(\mathbf{h}_i^k)\right]\right\}\\
		&\qquad* \left(1 - (\mathbf{h}_i^k)^2\right),
	\end{aligned}
	\label{eq:2}
\end{equation}
where $\mathbf{h}_i^k = \operatorname{tanh}(\operatorname{F}^k(\mathbf{x}_i;\theta))$ is the k-th bit of $\mathbf{h}_i$, $\operatorname{\sigma}(x) = \frac{1}{1 + e^{-x}}$ is the sigmoid function. The first term in Eq.~\eqref{eq:2} is the derivate of hash loss and the second is the derivate of $\operatorname{tanh}(\cdot)$.


\begin{figure}
	\centering
	\includegraphics[width=0.8\columnwidth]{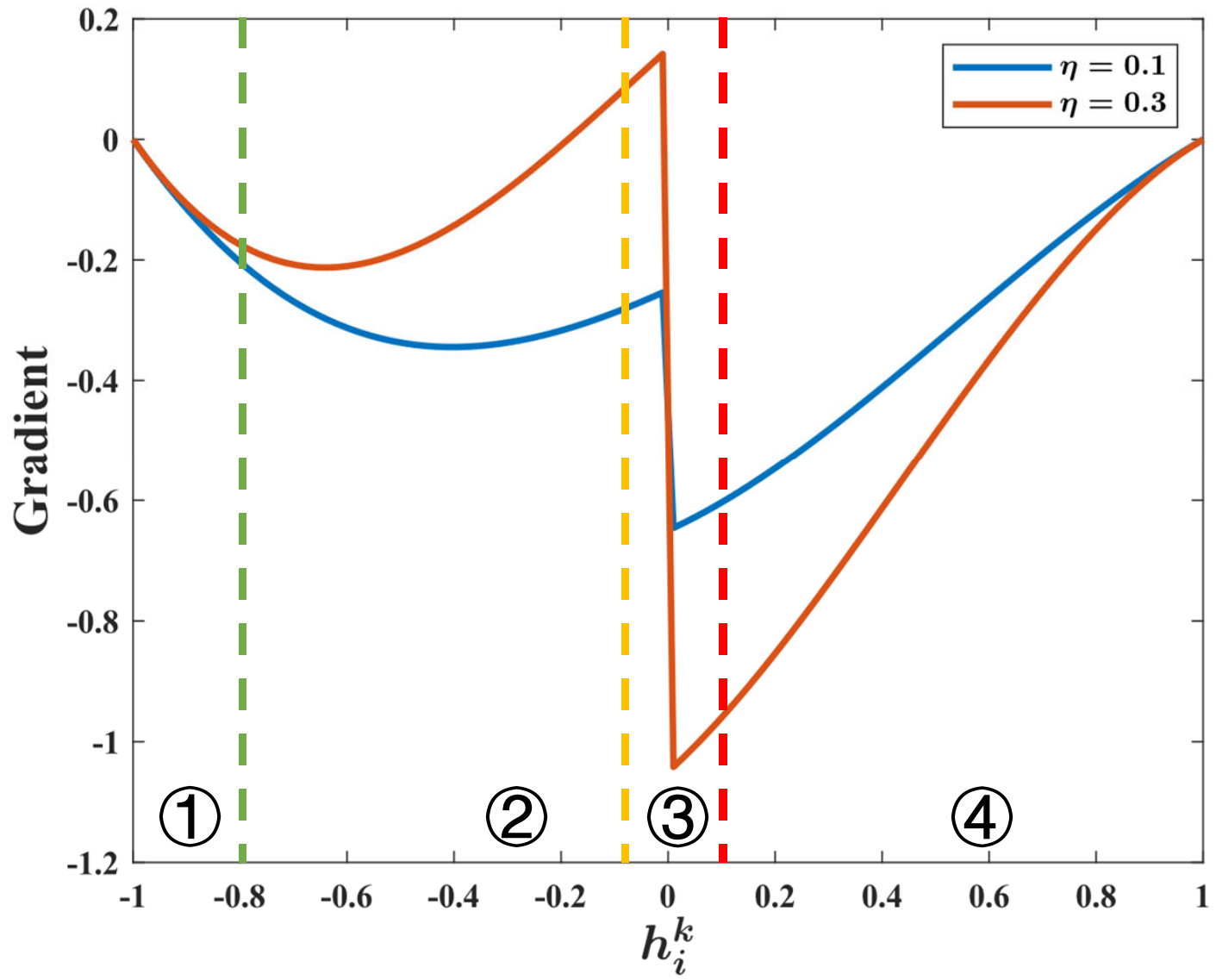}
	\caption{The visualization of hash loss's gradient. X-axis is split into 4 areas. In area \textcircled{1}, the saturated portion of $\operatorname{tanh}$ suppresses gradient; In area \textcircled{2}, the quantization loss is intensified and will prevent the bit from crossing zero point. If the hyper-parameter of quantization loss term is large, it may push the bit toward error direction; In area \textcircled{3}, there is a cliff around zero due to the change of the quantization loss's direction; In area \textcircled{4}, the gradient is decreasing as the distance between $\mathbf{h}_i^k$ and $\mathbf{h}_j^k$ achieves zero. Best viewed in color.}
	\label{fig:grad}
\end{figure}

\textbf{Why does DBP appear?} 
For simplicity, assume we have a pair of hash codes with one bit, $\mathbf{h}_j = 0.9, \mathbf{S}_{ij}=1, \eta=1$, and we visualize Eq.~\eqref{eq:2} w.r.t. $\mathbf{h}_i$, which is illustrated in Figure~\ref{fig:grad}.

The x-axis is divided into 4 areas. In area \textcircled{1}~(the left of green dashed line), the absolute value of gradient should be large because $\mathbf{h}_i^k$ is far away from $\mathbf{h}_j^k$ now. However, due to the second term in Eq.~\eqref{eq:2}, the saturated portion of $\operatorname{tanh}(\cdot)$ will suppress the gradient, especially when $\mathbf{h}_i^k$ is close to $-1$. Consequently, these bits can not be optimized. To minimize the loss, network tends to change others bits, which may lead to training instability. In area \textcircled{2}~(between green and yellow dashed line), the gradient seems to be decreased. This is because the quantization loss~(the second term in Eq.~\eqref{eq:1}) can not leverage the supervised information and only pushes the bit to $+1$ or $-1$~(depends on which side of axis it is on). When the bit is close to $0$ from the left side, the punishment of quantization loss will get intense and prevent it from crossing $0$ point. If $\eta$ is large, it may push the bit toward wrong direction, and the network can not be optimized correctly~(orange line). In area \textcircled{3}~(between yellow and red dashed line), there is a cliff around $0$. The reason is quantization loss changes the direction of punishment around zero point. If $\mathbf{h}_i^k$ approaches $0$ from the left side, the quantization loss will push it toward $-1$. Once it crosses zero point, the quantization pushed it toward $+1$ immediately. In area \textcircled{4}~(the right of read dashed line), the gradient is decreasing as the distance between $\mathbf{h}_i^k$ and $\mathbf{h}_j^k$ achieves zero.

The core reason leading to DBP is the saturated area in $\operatorname{tanh}(\cdot)$. From Eq.~\eqref{eq:2} and Figure~\ref{fig:grad}, if there is a continuous hash bit $\mathbf{h}_i^k = +1$ or $-1$, whatever $\mathbf{h}_j^k$ is, the gradient is always $0$. Furthermore, there are two reasons that may aggravate the problem. First, in Eq.~\eqref{eq:2}, the first term in braces shows the gradient of $\operatorname{F}^k(\mathbf{x}_i;\theta)$ is depended on the sum of the relations between $\mathbf{h}_i^k$ and $\mathbf{h}_j^k$. 
Limited by the noise in data and the capacity of CNN model, some bits may be optimized incorrectly. Once they get stuck in the saturated area, they can not escape from it. The other problem may aggravate DBP is the quantization loss. Existing quantization losses, e.g., $\text{L}_2$ quantization loss~\cite{li2015feature}, $\log\cosh$ quantization loss~\cite{zhu2016deep}, do not leverage the similarity information. They just push bit to $-1$ or $+1$~(depend on which side of axis it is on). If the parameter of quantization loss is selected improperly~(Figure~\ref{fig:grad}, orange line), it may push the bit toward wrong direction and lead to DBP. Furthermore, these dead bits will disturb the optimization process of others bits due to Eq.~\eqref{eq:2}, and lead to sub-optimal retrieval performance.




\section{Rescuing Deep Hashing}
To address DBP, we propose Rescuing Deep Hashing~(RDH), which consists of two components: Gradient Amplifier and Error-Aware Quantization Loss, as illustrated in Figure~\ref{fig:arch}.

\textbf{Gradient amplifier} aims to amplify the gradient of dead bits in saturated area. Its formulation is as following:
\begin{equation}
	\begin{aligned}
		\operatorname{GA}(\mathbf{g}_i^k) = \begin{cases}
			\mathbf{g}_i^k, & \vert\mathbf{h}_i^k\vert < \tau\\
			\alpha\cdot\mathbf{\mathbf{g}_i^k}, &\vert\mathbf{h}_i^k\vert \geq \tau, \operatorname{sign}(\mathbf{h}_i^k) = \operatorname{sign}(\mathbf{g}_i^k),
		\end{cases}
	\end{aligned}
	\label{eq:3}
\end{equation}
where $\operatorname{GA}(\cdot)$ is the gradient amplifier, $\mathbf{g}_i^k$ is the gradient of $\mathbf{h}_i^k$, $\alpha = \frac{1}{1 - \tau^2}$ is the amplification factor and $\tau \in [0, 1)$ is the threshold to determine whether to amplify gradient. 

During the forward-propagation phase, gradient amplifier collects continuous hash codes $\mathbf{h}$. At the back-propagation stage, it fetches gradients from later layers and takes out $\mathbf{h}$. If the direction of the gradient is equal to the bit, it means the bit need to be moved toward the opposite direction. Furthermore, the bit is in the saturated area of $\operatorname{tanh}(\cdot)$ if $\vert \mathbf{h}_i^k \vert \geq \tau$ and it can not escape from the area. Then gradient amplifier will magnify the gradient of it and help it move toward the correct direction.

\textbf{Error-aware quantization loss} is to solve the problem that ordinary quantization loss~\cite{li2015feature,zhu2016deep,li2020general} can not aware the correct direction of hash bits. The formulation is as following:
\begin{equation}
	\begin{aligned}
		\mathcal{L}_{\textbf{EAQ}} = \begin{cases}
			\Vert\mathbf{h}_{i/j}^k - \operatorname{sign}(\mathbf{h}_{i/j}^k)\Vert^2, &\text{if } (-1)^{\mathbf{S}_{ij}}\cdot\delta =-1\\
			0, &\text{else},
		\end{cases}
	\end{aligned}
	\label{eq:4}
\end{equation}
where $\delta = \operatorname{sign}(\mathbf{h}_i^k) \operatorname{sign}(\mathbf{h}_j^k)$.

\begin{table*}[ht]
	\centering
	\begin{tabular}{lrrrrrrrrrrrr}
		\toprule
		\multirow{2}{*}{Methods} & \multicolumn{4}{c}{CIFAR-10} & \multicolumn{4}{c}{MS-COCO} & \multicolumn{4}{c}{NUS-WIDE}\\
		& 24 bits & 32 bits & 48 bits & 64 bits & 24 bits & 32 bits & 48 bits & 64 bits & 24 bits & 32 bits & 48 bits & 64 bits\\
		\midrule
		DPSH & 0.7465 & 0.7467 & 0.7418 & 0.7466 & 0.6770 & 0.6946 & 0.7137 & 0.7109 & 0.8075 & 0.8137 & 0.8242 & 0.8265\\
		+RDH & \textbf{0.7621} & \textbf{0.7719} & \textbf{0.7748} & \textbf{0.7731} & \textbf{0.6986} & \textbf{0.7097} & \textbf{0.7189} & \textbf{0.7231} & \textbf{0.8250} & \textbf{0.8332} & \textbf{0.8387} & \textbf{0.8414}\\
		\midrule 
		DHN & 0.7426 & 0.7461 & 0.7449 & 0.7387 & 0.6894 & 0.6950 & 0.7081 & 0.7132 & 0.8198 & 0.8278 & 0.8326 & 0.8351\\
		+RDH & \textbf{0.7653} & \textbf{0.7704} & \textbf{0.7712} & \textbf{0.7651} & \textbf{0.7043} & \textbf{0.7168} & \textbf{0.7294} & \textbf{0.7354} & \textbf{0.8202} & \textbf{0.8307} & \textbf{0.8344} & \textbf{0.8370}\\
		\midrule 
		HashNet & 0.7477 & 0.7568 & 0.7630 & 0.7635 & 0.6877 & 0.6985 & 0.7100 & 0.7172 & 0.8176 & 0.8209 & 0.8303 & 0.8332\\
		+RDH & \textbf{0.7737} & \textbf{0.7804} & \textbf{0.7859} & \textbf{0.7866} & \textbf{0.7034} & \textbf{0.7167} & \textbf{0.7269} & \textbf{0.7302} & \textbf{0.8256} & \textbf{0.8311} & \textbf{0.8404} & \textbf{0.8455}\\
		\midrule
		DAGH & 0.7310 & 0.7218 & 0.7248 & 0.7269 & 0.6635 & 0.6720 & 0.6854 & 0.6899 & 0.8185 & 0.8254 & 0.8313 & 0.8354\\
		+RDH & \textbf{0.7561} & \textbf{0.7579} & \textbf{0.7622} & \textbf{0.7559} & \textbf{0.6897} & \textbf{0.7051} & \textbf{0.7084} & \textbf{0.7131} & \textbf{0.8233} & \textbf{0.8296} & \textbf{0.8374} & \textbf{0.8392}\\
		\midrule 
		DSDH & 0.7584 & 0.7611 & 0.7740 & 0.7703 & 0.7199 & 0.7474 & 0.7728 & 0.7805 & 0.8152 & 0.8206 & 0.8279 & 0.8326\\
		+RDH & \textbf{0.7767} & \textbf{0.7886} & \textbf{0.7845} & \textbf{0.7887} & \textbf{0.7285} & \textbf{0.7539} & \textbf{0.7764} & \textbf{0.7862} & \textbf{0.8185} & \textbf{0.8297} & \textbf{0.8416} & \textbf{0.8461}\\
		\bottomrule
	\end{tabular}
	\caption{Comparison of MAP with different bits on CIFAR-10, MS-COCO and NUS-WIDE. The best accuracy is shown in boldface.}
	\label{tab:1}
\end{table*}

Traditional quantization loss just pushes the continuous bit toward $-1$, if the bit is on the left of $0$; $+1$ otherwise. It does not leverage the similarity information and may push the bit toward the wrong direction. Error-aware quantization loss chooses bits that satisfy 1). $\mathbf{S}_{ij}=1$ and $\operatorname{sign}(\mathbf{h}_i^k) = \operatorname{sign}(\mathbf{h}_j^k)$; 2). $\mathbf{S}_{ij}=0$ and $\operatorname{sign}(\mathbf{h}_i^k) \neq \operatorname{sign}(\mathbf{h}_j^k)$. For others bits, we ignore and delegate them to the similarity preserving loss to optimize them, because the number of correct and wrong bits is imbalance and dynamic during training. Optimizing them will introduce extra hyper-parameter and increase the difficulty of tuning.

Due to the flexibility of gradient amplifier and error-aware quantization loss, they can be seamlessly assembled with existing similarity preserving loss in hashing methods~\cite{li2015feature,cao2017hashnet,chen2019deep}, we denote these losses as  $\mathcal{L}_{\text{sim}}$. Finally, the overall objective function is 
\begin{equation}
	\begin{aligned}
		\min_\theta \mathcal{L} = \min_\theta (\mathcal{L}_{\text{sim}} + \eta\mathcal{L}_{\text{EQA}}).
	\end{aligned}
	\label{eq:5}
\end{equation}

The learning algorithm is summarized in Algorithm~\ref{alg:1}.

\begin{figure*}[ht]
	\centering
	\subfloat{\includegraphics[width=2in]{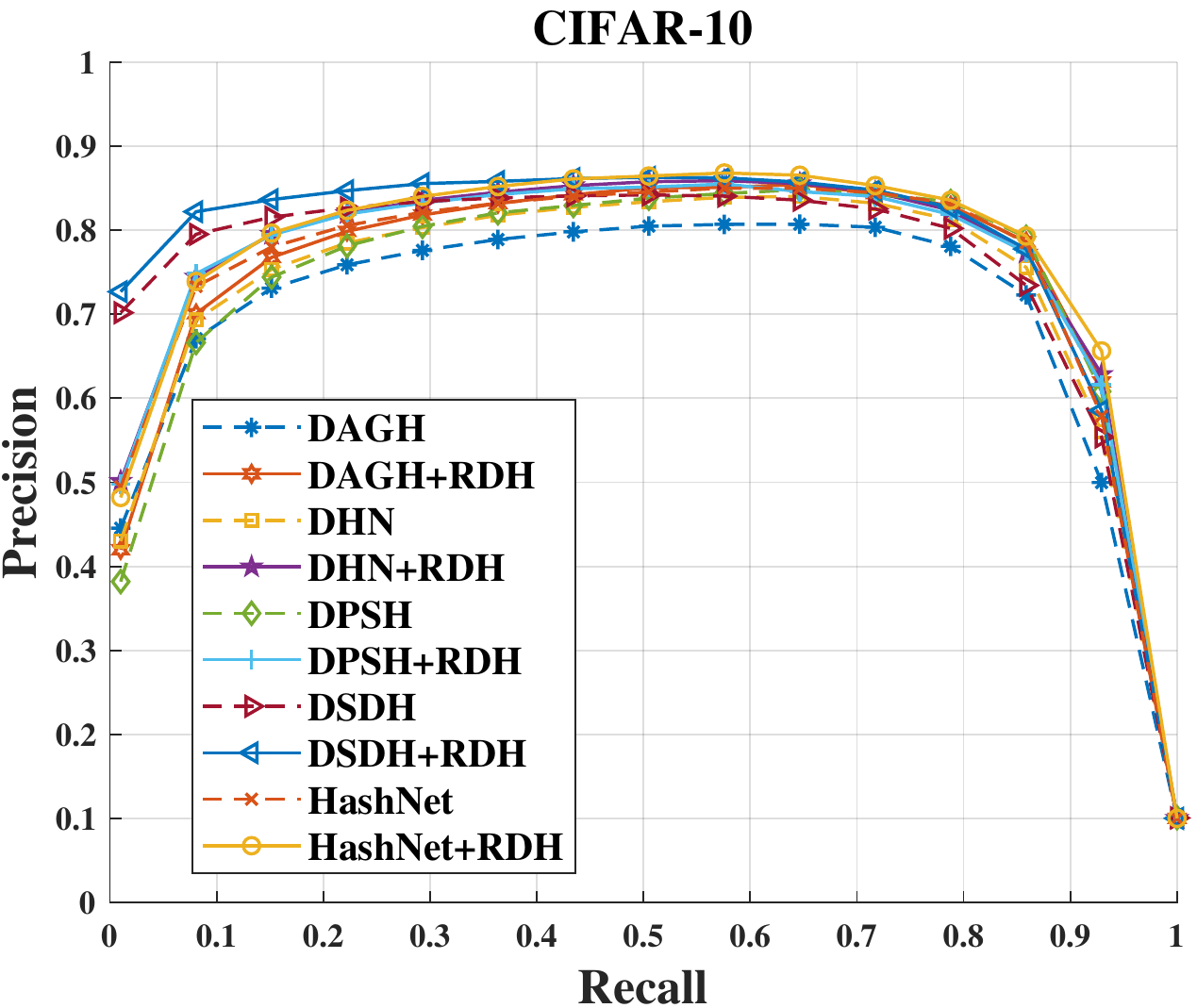}}
	\quad
	\subfloat{\includegraphics[width=2in]{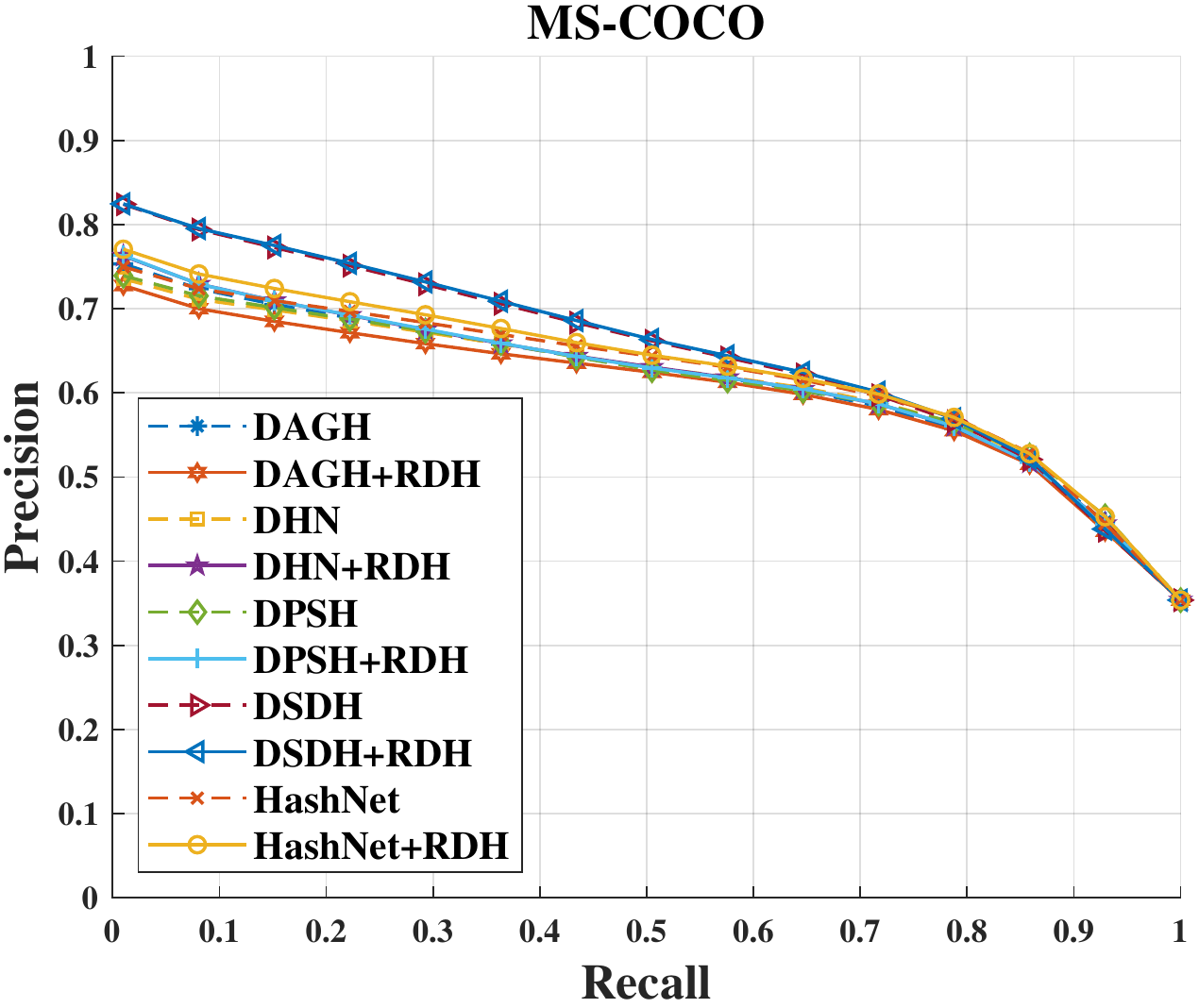}}
	\quad
	\subfloat{\includegraphics[width=2in]{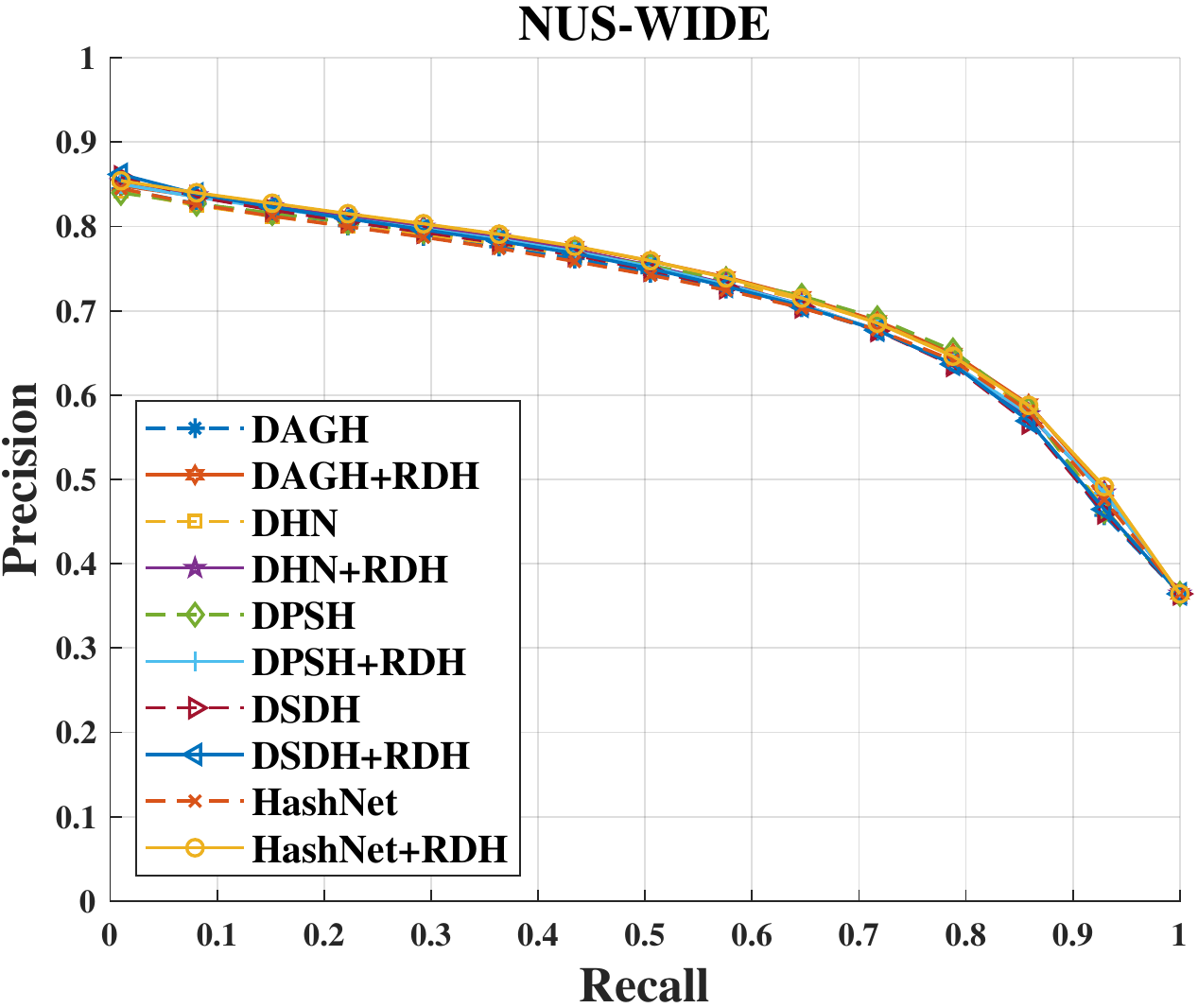}}
	\caption{Precision-Recall curves with a code length of 64 on three datasets. Best viewed in color.}
	\label{fig:prcurve}
\end{figure*}

\begin{algorithm}[tb]
	\caption{The learning algorithm for RDH}
	\label{alg:1}
	\textbf{Input}: training images $\mathbb{X}$; Similarity matrix $\mathbf{S}$; Threshold $\tau$; Hyper-parameter $\eta$\\
	\textbf{Output}: CNN model $\operatorname{F}(\mathbf{x};\theta)$
	\begin{algorithmic}[1] 
		\STATE Initialize a new CNN model and replace the last fully connected layer with a new $K$-dimension fully connected layer followed by $\operatorname{tanh}(\cdot)$ activation function.
		\WHILE{Not convergence or not reach maximum iterations}
		\STATE Forward propagate and get continuous hash codes $\mathbf{h}$.
		\STATE Store $\mathbf{h}$ for gradient amplification.
		\STATE Recognize bits need to be quantified according to Eq.~\eqref{eq:4}.
		\STATE Calculate the loss by Eq.~\eqref{eq:5}.		
		\STATE Backward propagate to obtain the derivate $\partial \mathcal{L}/\partial \mathbf{h}_i^k$.
		\STATE Recognize dead bits according to Eq.~\eqref{eq:3}.
		\STATE Amplify gradients through Gradient Amplifier according to Eq.~\eqref{eq:3}.
		\STATE Update model parameters $\theta$.
		\ENDWHILE
		\STATE \textbf{return} CNN model $\operatorname{F}(\mathbf{x};\theta)$
	\end{algorithmic}
\end{algorithm}

\section{Experiments}
\subsection{Datasets}
The experiments of RDH are conducted on three widely used datasets: \textbf{CIFAR-10}~\cite{krizhevsky2009learning}, \textbf{MS-COCO}~\cite{lin2014microsoft}, and \textbf{NUS-WIDE}~\cite{chua2009nus}.

\begin{itemize}
	\item \textbf{CIFAR-10}\footnote{http://www.cs.toronto.edu/$\sim$kriz/cifar.html} is a single-label dataset, containing 60,000 color images with $32 \times 32$ resolution, belonging to 10 classes. Each class has 6,000 images. Following~\cite{li2015feature,wu2019deep,zhao2020asymmetric,li2020general}, we randomly sample 1,000 images~(100 images per class) as the query set, and the rest are used to form the gallery set. Then we randomly select 5,000 images~(500 images per class) from the gallery as the training set.
	\item
	\textbf{MS-COCO}\footnote{https://cocodataset.org/} is a multi-label dataset. It contains 82,783 training images and 40,504 validation images, which belong to 80 classes. We obtain 12,2218 images by combining the training and validation images and pruning images without category annotation. Following \cite{cao2017hashnet,jiang2018asymmetric}, we randomly sample 5,000 images as the query set, and the rest images are used as the gallery. Furthermore, we random select 10,000 images from the gallery as the training points.
	\item
	\textbf{NUS-WIDE}\footnote{\text{https://lms.comp.nus.edu.sg/wp-content/uploads/2019/research}\\\text{\qquad/nuswide/NUS-WIDE.html}} is a multi-label dataset, which contains 269,648 images from 81 classes. Following \cite{lai2015simultaneous,jiang2018asymmetric,zhao2020asymmetric}, we use a subset of original dataset which associates with the 21 most frequent classes to conduct our experiment. Specially, we randomly select 2,100 images (100 images per class) as the query set. From the rest images, we randomly choose 10,500 images (500 images per class) to make up the training set.
\end{itemize}

\subsection{Evaluation Methodology}
Following~\cite{li2015feature,wu2019deep,zhao2020asymmetric,li2020general}, we adopt the Mean Average Precision~(MAP) and Precision-Recall~(PR) curves to evaluate the retrieval performance. Particularly, for MS-COCO and NUS-WIDE, the MAP is calculated based on the Top-5K returned samples; for ImageNet, it is calculated within the Top-1K returned neighbors. For multi-label dataset, e.g. MS-COCO, two images are considered to be similar if they share at least one common label. All experiments are conducted 5 times, and we report the average results of them.

\subsection{Experimental Details}
As a plug-in module, RDH can be adopted with various methods trained with different models and loss functions. We selected some representative deep hashing methods, e.g.  DPSH~\cite{li2015feature}, DHN~\cite{zhu2016deep}, HashNet~\cite{cao2017hashnet}, DAGH~\cite{chen2019deep}, DSDH~\cite{li2020general} as the baseline methods. All of these methods are implemented on PyTorch~\cite{paszke2019pytorch}. 

\begin{itemize}
	\item \textbf{Network Design}. For a fair comparison with other state-of-the-art methods, we retrain all the baseline methods, employing AlexNet~\cite{krizhevsky2017imagenet} pretrained on ImageNet~\cite{deng2009imagenet} as the backbone. The last fully connected layer is removed, and replaced with a new one, where the dimension of the outputs is the hash code length. 
	\item \textbf{Training Details} In RDH, SGD is utilized as the optimizer with $1\mathrm{e}{-5}$ weight decay. The initial learning rate is set to $1\mathrm{e}{-2}$. Cosine annealing the learning rate scheduler~\cite{loshchilov2016sgdr} is leveraged to gradually reduce learning rate to zero. The batch size is set to 128. 
	\item \textbf{Network Parameters} We set $\tau=0.99$. $\eta$ is set to $1$ for CIFAR-10 and $0.1$ for the others. 
\end{itemize}

All the experiments are conducted on a single NIVDIA RTX 2080ti GPU.

\subsection{Accuracy Comparison}
In this section, we compare the performance between with and without RDH on CIFAR-10, MS-COCO and NUS-WIDE. Table~\ref{tab:1} shows the MAP of different methods with various hash code lengths. ``+RDH'' represents training model with the gradient amplifier and the error-aware quantization loss. 

Figure~\ref{fig:prcurve} illustrate the Precision-Recall curves. The results indicates RDH has ability to improve the retrieval performance of the baseline methods and can be used as a plug-and-play component to fit most existing hashing methods. Compared with the baseline methods, RDH outperforms them across all datasets. For example, our proposed method improves the average MAP by 3.29\% and 2.33\% on CIFAR-10 and MS-COCO respectively. And we can observe the similar results on NUS-WIDE. In addition, on CIFAR-10 dataset, we observe that RDH methods with 24 bits are even better than their baseline methods with 64 bits. It indicates the benefit that dead bits are rescued from the saturated area by our proposed gradient amplifier and error-aware quantization loss have ability to rescue dead bits from the saturated area. Consequently, the retrieval performance is improved significantly.

\subsection{Effectiveness of Gradient Amplifier and Error-Aware Quantization Loss}
\begin{figure}[ht]
	\centering
	\includegraphics[width=0.75\columnwidth]{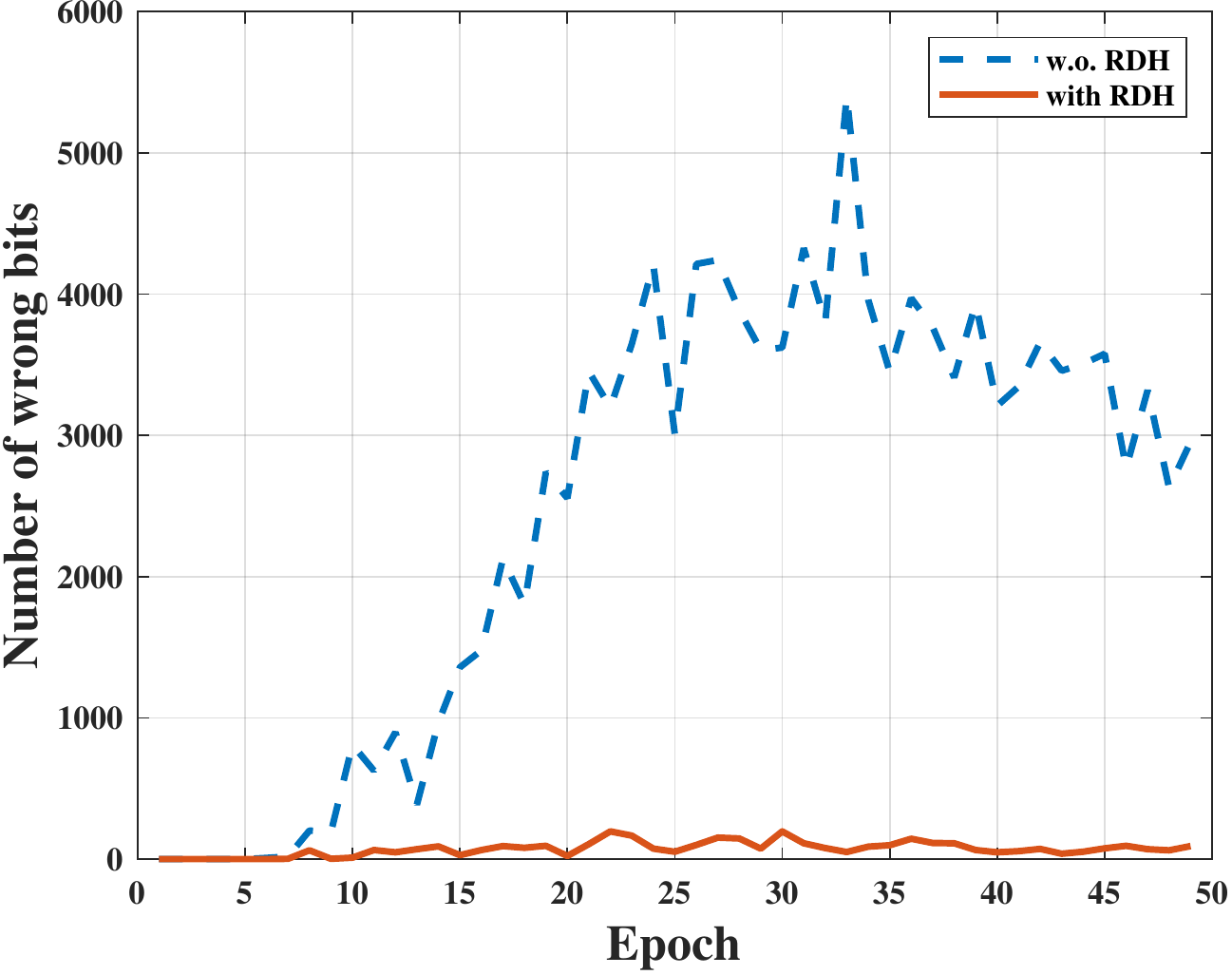}
	\caption{The number of dead bits with/without gradient amplifier and error-aware quantization loss. Best viewed in color.}
	\label{fig:rdh}
\end{figure}

We count the dead bits with/without gradient amplifier and error-aware quantization loss during the training phase. The results are shown in Figure~\ref{fig:rdh}. 

In Figure~\ref{fig:rdh}, blue line represents the results without our proposed gradient amplifier and error-aware quantization, where dead bits are increasing during the training stage and the number of them is kept at a high level. This is mainly because some bits ``incautiously'' arrive in the saturated area and hard to escape from it. As a result the existence of our proposed `Dead Bits Problem'' can be confirmed.

Nonetheless, with the help of gradient amplifier and error-aware quantization, DBP is mitigated significantly, which is shown by the orange line in Figure~\ref{fig:rdh}. Specifically, the number of dead bits is decreased significantly and stays below $200$ during the training time. It means many dead bits can escape from the saturated area and have the opportunity to be optimized correctly. Consequently, only few bits get stuck in saturated area mainly due to the noise of dataset or the capacity of CNN model.

\subsection{Ablation Study}
To analyze the effectiveness of gradient amplifier and error-aware quantization loss, we design some variants of our proposed algorithm, and show some empirical analysis. We compare our algorithm with the following baseline: \textbf{(1)~Without gradients amplifier~(w.o. GA)}. The gradients amplifier is removed in this baseline. \textbf{(2)~Without error-aware quantization loss~(w.o. EAQL)}. We remove the error-aware quantization loss.

\begin{table}[htb]
	\centering
	\begin{tabular}{lcccc}
		\toprule
		Methods & 24 bits & 32 bits & 48 bits & 64 bits\\
		\midrule
		HashNet+RDH & \textbf{0.7737} & \textbf{0.7804} & \textbf{0.7859} & \textbf{0.7866}\\
		\midrule
		w.o. GA & 0.7612 & 0.7649 & 0.7686 & 0.7601 \\
		w.o. EAQL & 0.7691 & 0.7679 & 0.7793 & 0.7696\\
		HashNet & 0.7477 & 0.7573 & 0.7630 & 0.7635 \\
		\bottomrule
	\end{tabular}
	\caption{MAP results by using different variants on CIFAR-10. We employ HashNet as backbone. The best accuracy is shown in boldface.}
	\label{tab:ablation-study}
\end{table}

Table~\ref{tab:ablation-study} shows the performance of there variants. We can observe that the retrieval performance drops dramatically when modifying the structure of proposed algorithm. Specifically, when we remove gradient amplifier, dead bits can not be pushed away from saturated area, and the MAP decreases by ~2\%. When removing error-aware quantization loss, some bits may hard to cross the zero point, leading to the degradation of retrieval performance.

\subsection{Sensitivity Analysis}
\begin{figure}[htb]
	\centering
	\subfloat{\includegraphics[width=1.5in]{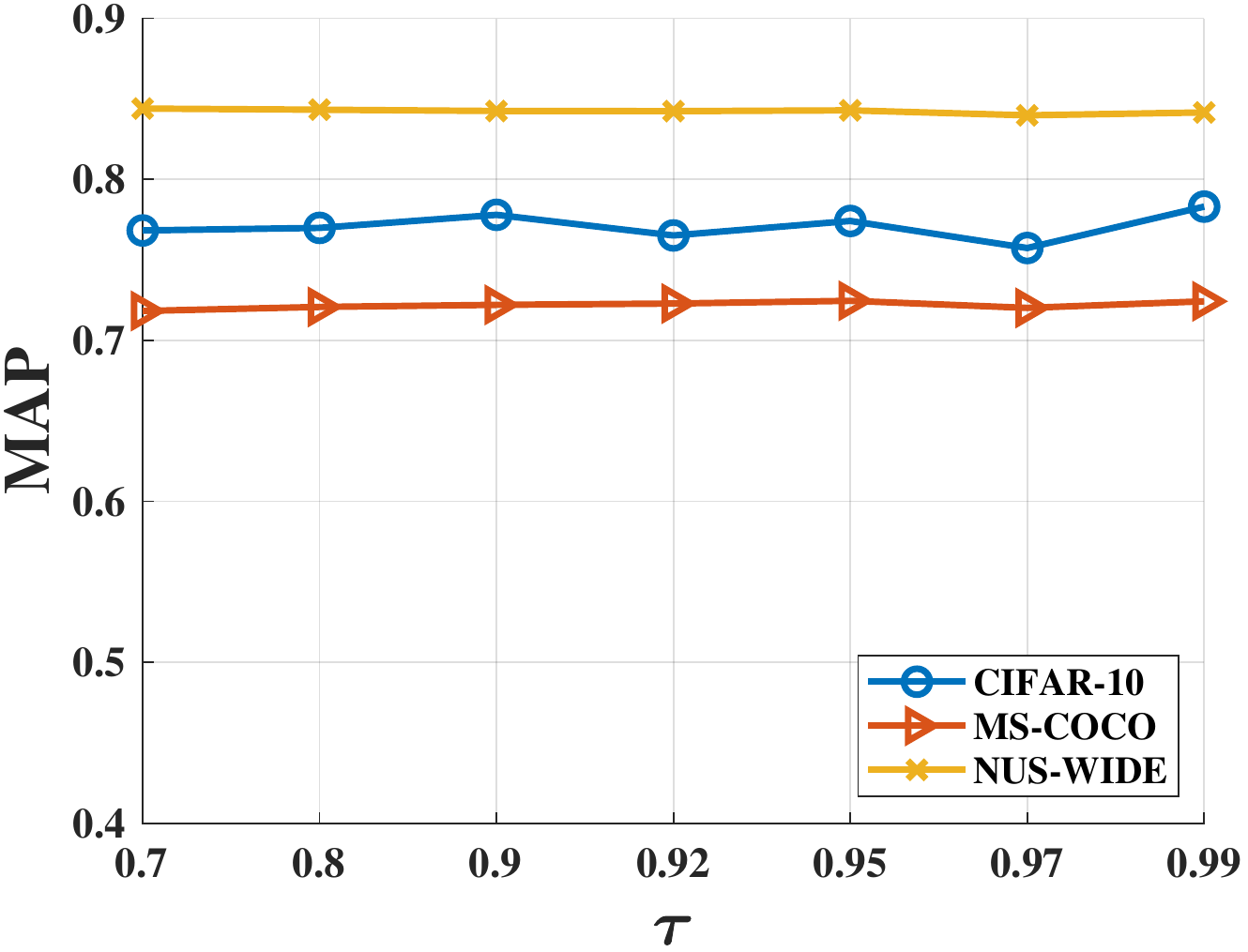}}
	\quad
	\subfloat{\includegraphics[width=1.5in]{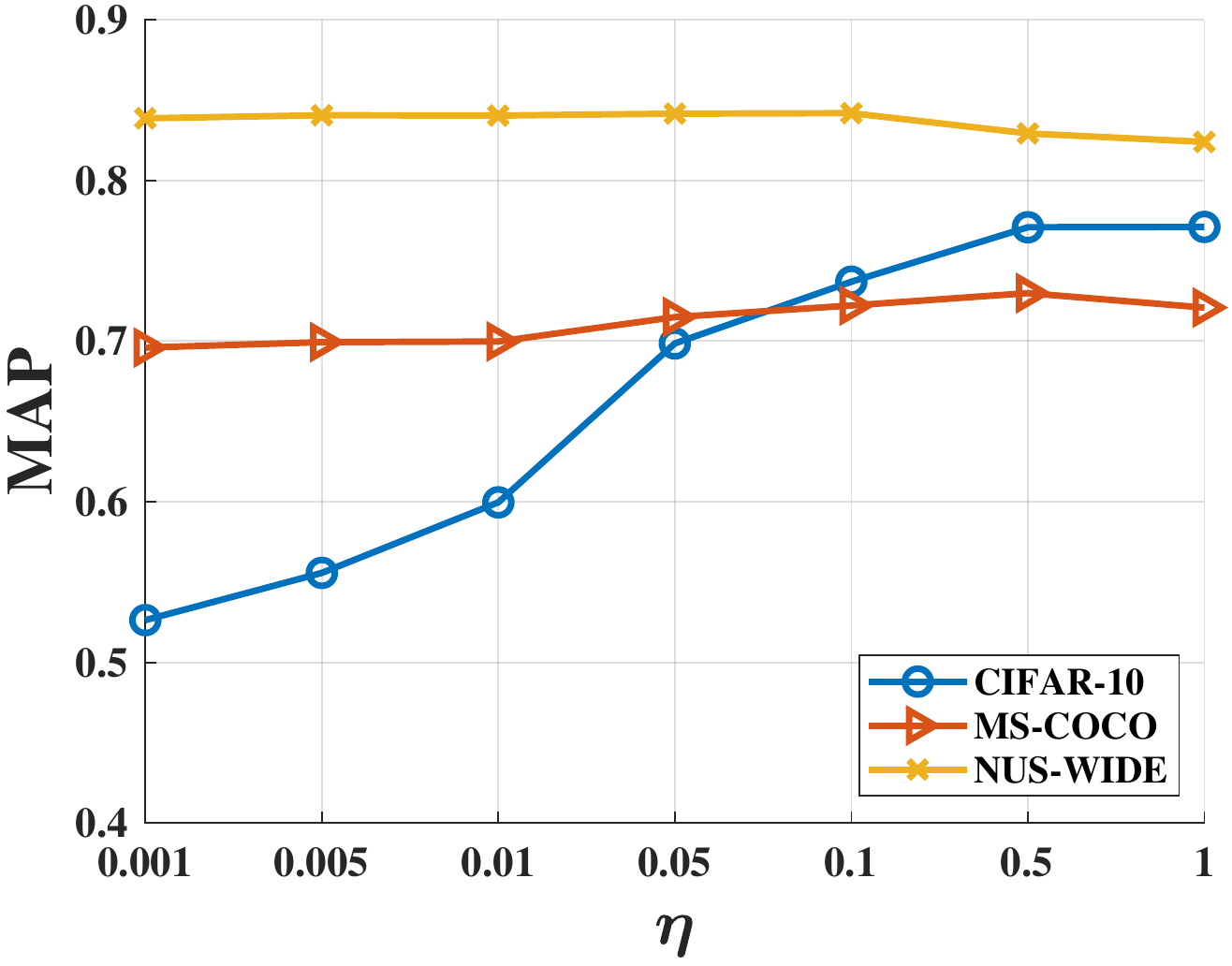}}
	\caption{Sensitivity on three datasets with different $\tau$ and $\eta$. Best viewed in color.}
	\label{fig:sensitivity}
\end{figure}

To investigate the sensitivity of the hyper-parameters $\tau$ and $\eta$, we further conduct experiments under different values of $\tau$ and $\eta$, the retrieval performance is illustrated in Figure~\ref{fig:sensitivity}. $\tau$ is used to control the threshold of range gradients amplifier effects and $\eta$ is leveraged to balance the trade-off between original loss and EAQ loss. 

From Figure~\ref{fig:sensitivity}, we observe that our algorithm is not sensitive to $\tau$. This is mainly because gradient amplifier can amplify gradients of dead bits adaptively according to Eq.~\eqref{eq:3}. 

For $\eta$, it is not sensitive on multi-label datasets, e.g., MS-COCO and NUS-WIDE. However, on CIFAR-10, the retrieval performance decreases when $\eta$ is set to a small value. This is mainly because the resolution of images in CIFAR-10 is low and leads to high uncertainty of bits, i.e., the bits will be oscillatory during training time. Large $\eta$ can help stabilize these bits and improve the retrieval performance. 

\section{Conclusion}
In this paper, we have characterized a serious problem that caused by the saturated area in activation function, e.g. $\operatorname{sigmoid}(\cdot)$ or $\operatorname{tanh}(\cdot)$ and the traditional quantization loss, called DBP. To address this issue, we have proposed a gradient amplifier to detect and rescue the dead bits. Furthermore, an error-aware quantization loss is proposed to alleviate DBP. Extensive experiments have  demonstrated that the proposed method can significantly decrease the number of dead bits and improve the performance of the baseline methods.

\bibliographystyle{named}
\bibliography{ijcai21}

\begin{thebibliography}{}

\bibitem[\protect\citeauthoryear{Cao \bgroup \em et al.\egroup
  }{2017}]{cao2017hashnet}
Zhangjie Cao, Mingsheng Long, Jianmin Wang, and Philip~S Yu.
\newblock Hashnet: Deep learning to hash by continuation.
\newblock In {\em ICCV}, pages 5608--5617, 2017.

\bibitem[\protect\citeauthoryear{Cao \bgroup \em et al.\egroup
  }{2018}]{cao2018deep}
Yue Cao, Mingsheng Long, Bin Liu, and Jianmin Wang.
\newblock Deep cauchy hashing for hamming space retrieval.
\newblock In {\em Proceedings of the IEEE Conference on Computer Vision and
  Pattern Recognition}, pages 1229--1237, 2018.

\bibitem[\protect\citeauthoryear{Chen \bgroup \em et al.\egroup
  }{2019}]{chen2019deep}
Yudong Chen, Zhihui Lai, Yujuan Ding, Kaiyi Lin, and Wai~Keung Wong.
\newblock Deep supervised hashing with anchor graph.
\newblock In {\em ICCV}, pages 9796--9804, 2019.

\bibitem[\protect\citeauthoryear{Chua \bgroup \em et al.\egroup
  }{2009}]{chua2009nus}
Tat-Seng Chua, Jinhui Tang, Richang Hong, Haojie Li, Zhiping Luo, and Yantao
  Zheng.
\newblock Nus-wide: a real-world web image database from national university of
  singapore.
\newblock In {\em ICIVR}, pages 1--9, 2009.

\bibitem[\protect\citeauthoryear{Deng \bgroup \em et al.\egroup
  }{2009}]{deng2009imagenet}
Jia Deng, Wei Dong, Richard Socher, Li-Jia Li, Kai Li, and Li~Fei-Fei.
\newblock Imagenet: A large-scale hierarchical image database.
\newblock In {\em CVPR}, pages 248--255. Ieee, 2009.

\bibitem[\protect\citeauthoryear{Fan \bgroup \em et al.\egroup
  }{2020}]{fan20deep}
Lixin Fan, Kam~Woh Ng, Ce~Ju, Tianyu Zhang, and Chee~Seng Chan.
\newblock Deep polarized network for supervised learning of accurate binary
  hashing codes.
\newblock In {\em IJCAI}, pages 825--831, 2020.

\bibitem[\protect\citeauthoryear{Fu \bgroup \em et al.\egroup
  }{2020}]{fu2020deep}
Chaoyou Fu, Guoli Wang, Xiang Wu, Qian Zhang, and Ran He.
\newblock Deep momentum uncertainty hashing.
\newblock {\em arXiv preprint arXiv:2009.08012}, 2020.

\bibitem[\protect\citeauthoryear{Gionis \bgroup \em et al.\egroup
  }{1999}]{gionis1999similarity}
Aristides Gionis, Piotr Indyk, Rajeev Motwani, et~al.
\newblock Similarity search in high dimensions via hashing.
\newblock In {\em VLDB}, volume~99, pages 518--529, 1999.

\bibitem[\protect\citeauthoryear{Gong \bgroup \em et al.\egroup
  }{2012}]{gong2012iterative}
Yunchao Gong, Svetlana Lazebnik, Albert Gordo, and Florent Perronnin.
\newblock Iterative quantization: A procrustean approach to learning binary
  codes for large-scale image retrieval.
\newblock {\em TPAMI}, 35(12):2916--2929, 2012.

\bibitem[\protect\citeauthoryear{Gui \bgroup \em et al.\egroup
  }{2017}]{gui2017fast}
Jie Gui, Tongliang Liu, Zhenan Sun, Dacheng Tao, and Tieniu Tan.
\newblock Fast supervised discrete hashing.
\newblock {\em IEEE transactions on pattern analysis and machine intelligence},
  40(2):490--496, 2017.

\bibitem[\protect\citeauthoryear{Jiang and Li}{2017}]{jiang2017deep}
Qing-Yuan Jiang and Wu-Jun Li.
\newblock Deep cross-modal hashing.
\newblock In {\em CVPR}, pages 3232--3240, 2017.

\bibitem[\protect\citeauthoryear{Jiang and Li}{2018}]{jiang2018asymmetric}
Qing-Yuan Jiang and Wu-Jun Li.
\newblock Asymmetric deep supervised hashing.
\newblock In {\em AAAI}, volume~32, 2018.

\bibitem[\protect\citeauthoryear{Krizhevsky \bgroup \em et al.\egroup
  }{2009}]{krizhevsky2009learning}
Alex Krizhevsky, Geoffrey Hinton, et~al.
\newblock Learning multiple layers of features from tiny images.
\newblock 2009.

\bibitem[\protect\citeauthoryear{Krizhevsky \bgroup \em et al.\egroup
  }{2017}]{krizhevsky2017imagenet}
Alex Krizhevsky, Ilya Sutskever, and Geoffrey~E Hinton.
\newblock Imagenet classification with deep convolutional neural networks.
\newblock {\em Communications of the ACM}, 60(6):84--90, 2017.

\bibitem[\protect\citeauthoryear{Lai \bgroup \em et al.\egroup
  }{2015}]{lai2015simultaneous}
Hanjiang Lai, Yan Pan, Ye~Liu, and Shuicheng Yan.
\newblock Simultaneous feature learning and hash coding with deep neural
  networks.
\newblock In {\em CVPR}, pages 3270--3278, 2015.

\bibitem[\protect\citeauthoryear{Lai \bgroup \em et al.\egroup
  }{2016}]{lai2016instance}
Hanjiang Lai, Pan Yan, Xiangbo Shu, Yunchao Wei, and Shuicheng Yan.
\newblock Instance-aware hashing for multi-label image retrieval.
\newblock {\em TIP}, 25(6):2469--2479, 2016.

\bibitem[\protect\citeauthoryear{Li \bgroup \em et al.\egroup
  }{2015}]{li2015feature}
Wu-Jun Li, Sheng Wang, and Wang-Cheng Kang.
\newblock Feature learning based deep supervised hashing with pairwise labels.
\newblock {\em arXiv preprint arXiv:1511.03855}, 2015.

\bibitem[\protect\citeauthoryear{Li \bgroup \em et al.\egroup
  }{2020}]{li2020general}
Qi~Li, Zhenan Sun, Ran He, and Tieniu Tan.
\newblock A general framework for deep supervised discrete hashing.
\newblock {\em IJCV}, 128(8):2204--2222, 2020.

\bibitem[\protect\citeauthoryear{Lin \bgroup \em et al.\egroup
  }{2014}]{lin2014microsoft}
Tsung-Yi Lin, Michael Maire, Serge Belongie, James Hays, Pietro Perona, Deva
  Ramanan, Piotr Doll{\'a}r, and C~Lawrence Zitnick.
\newblock Microsoft coco: Common objects in context.
\newblock In {\em ECCV}, pages 740--755. Springer, 2014.

\bibitem[\protect\citeauthoryear{Loshchilov and
  Hutter}{2016}]{loshchilov2016sgdr}
Ilya Loshchilov and Frank Hutter.
\newblock Sgdr: Stochastic gradient descent with warm restarts.
\newblock {\em arXiv preprint arXiv:1608.03983}, 2016.

\bibitem[\protect\citeauthoryear{Paszke \bgroup \em et al.\egroup
  }{2019}]{paszke2019pytorch}
Adam Paszke, Sam Gross, Francisco Massa, Adam Lerer, James Bradbury, Gregory
  Chanan, Trevor Killeen, Zeming Lin, Natalia Gimelshein, Luca Antiga, et~al.
\newblock Pytorch: An imperative style, high-performance deep learning library.
\newblock In {\em NeurIPS}, pages 8026--8037, 2019.

\bibitem[\protect\citeauthoryear{Shen \bgroup \em et al.\egroup
  }{2015}]{shen2015supervised}
Fumin Shen, Chunhua Shen, Wei Liu, and Heng Tao~Shen.
\newblock Supervised discrete hashing.
\newblock In {\em CVPR}, pages 37--45, 2015.

\bibitem[\protect\citeauthoryear{Shen \bgroup \em et al.\egroup
  }{2017}]{shen2017deep}
Fumin Shen, Xin Gao, Li~Liu, Yang Yang, and Heng~Tao Shen.
\newblock Deep asymmetric pairwise hashing.
\newblock In {\em ACM MM}, pages 1522--1530, 2017.

\bibitem[\protect\citeauthoryear{Wu \bgroup \em et al.\egroup
  }{2019}]{wu2019deep}
Dayan Wu, Qi~Dai, Jing Liu, Bo~Li, and Weiping Wang.
\newblock Deep incremental hashing network for efficient image retrieval.
\newblock In {\em CVPR}, pages 9069--9077, 2019.

\bibitem[\protect\citeauthoryear{Xia \bgroup \em et al.\egroup
  }{2014}]{xia2014supervised}
Rongkai Xia, Yan Pan, Hanjiang Lai, Cong Liu, and Shuicheng Yan.
\newblock Supervised hashing for image retrieval via image representation
  learning.
\newblock In {\em AAAI}, volume~28, 2014.

\bibitem[\protect\citeauthoryear{Yuan \bgroup \em et al.\egroup
  }{2020}]{yuan2020central}
Li~Yuan, Tao Wang, Xiaopeng Zhang, Francis~EH Tay, Zequn Jie, Wei Liu, and
  Jiashi Feng.
\newblock Central similarity quantization for efficient image and video
  retrieval.
\newblock In {\em CVPR}, pages 3083--3092, 2020.

\bibitem[\protect\citeauthoryear{Zhao \bgroup \em et al.\egroup
  }{2015}]{zhao2015deep}
Fang Zhao, Yongzhen Huang, Liang Wang, and Tieniu Tan.
\newblock Deep semantic ranking based hashing for multi-label image retrieval.
\newblock In {\em CVPR}, pages 1556--1564, 2015.

\bibitem[\protect\citeauthoryear{Zhao \bgroup \em et al.\egroup
  }{2020}]{zhao2020asymmetric}
Shu Zhao, Dayan Wu, Wanqian Zhang, Yu~Zhou, Bo~Li, and Weiping Wang.
\newblock Asymmetric deep hashing for efficient hash code compression.
\newblock In {\em ACM MM}, pages 763--771, 2020.

\bibitem[\protect\citeauthoryear{Zhu \bgroup \em et al.\egroup
  }{2016}]{zhu2016deep}
Han Zhu, Mingsheng Long, Jianmin Wang, and Yue Cao.
\newblock Deep hashing network for efficient similarity retrieval.
\newblock In {\em AAAI}, volume~30, 2016.

\end{thebibliography}

\end{document}